\documentclass[acmsmall]{acmart}

\setcopyright{none}
\renewcommand\footnotetextcopyrightpermission[1]{}
\renewcommand{\keywordsname}{\textbf{Keywords}}

\makeatletter
\def\@mkbibcitation{}
\def\@acmArticle{}
\def\@acmVolume{}
\def\@acmNumber{}
\def\@acmPubDate{}
\def\@formatdoi#1{}

\fancypagestyle{firstpagestyle}{%
  \fancyhf{}%
}

\AtBeginDocument{%
  \thispagestyle{firstpagestyle}
  \let\old@footnotetext\@footnotetext
  \renewcommand{\@footnotetext}[1]{}
}
\makeatother

\AtBeginDocument{%
  }

\graphicspath{{./images/}} 
\usepackage{graphicx} 
\usepackage{caption}  
\usepackage{longtable}
\usepackage{booktabs}
\usepackage{array}       
\usepackage{tabularx}
\usepackage{booktabs}

\usepackage[T1]{fontenc}
\usepackage{hyperref}
\usepackage{listings}
\usepackage{enumitem}

\definecolor{codegreen}{rgb}{0,0.6,0}
\definecolor{codegray}{rgb}{0.5,0.5,0.5}
\definecolor{codepurple}{rgb}{0.58,0,0.82}
\definecolor{backcolour}{rgb}{0.95,0.95,0.92}

\setlist[itemize]{leftmargin=*}
\setlist[enumerate]{leftmargin=*}

\lstdefinestyle{mystyle}{
  backgroundcolor=\color{backcolour}, 
  numberstyle=\tiny\color{codegray},
  basicstyle=\ttfamily\footnotesize,
  breakatwhitespace=false,         
  breaklines=true,                 
  captionpos=b,                    
  keepspaces=true,                 
  numbers=left,                    
  numbersep=5pt,                  
  showspaces=false,                
  showstringspaces=false,
  showtabs=false,                  
  tabsize=2,
 stringstyle=\ttfamily,
  commentstyle=\ttfamily,
  morecomment=[s]{"""}{"""},
  morestring=[s]{"""}{"""}
}

\begin{document}

\title{Crowdsourcing-Based Knowledge Graph Construction for Drug Side Effects Using Large Language Models with an Application on Semaglutide}

\author{Zhijie Duan}
\authornote{Both authors contributed equally to this research.}
\email{zhijie.duan@pennmedicine.upenn.edu}
\author{Kai Wei}
\authornotemark[1]
\email{weikai@umich.edu}
\affiliation{%
  \institution{University of Pennsylvania}
  \city{Philadelphia}
  \state{PA}
  \institution{University of Michigan}
  \city{Ann Arbor}
  \state{MI}
  \country{USA}
}

\author{Zhaoqian Xue}
\affiliation{%
  \institution{Georgetown University}
  \city{Washington DC}
  \country{USA}}
\email{zx136@georgetown.edu}

\author{Jiayan Zhou}
\affiliation{%
 \institution{Stanford University School of Medicine}
  \city{Stanford}
  \state{CA}
  \country{USA}}
\email{jyzhou@stanford.edu}

\author{Shu Yang}
\affiliation{%
  \institution{University of Pennsylvania}
  \city{Philadelphia}
  \state{PA}
  \country{USA}}
\email{shu.yang@pennmedicine.upenn.edu}

\author{Siyuan Ma}
\affiliation{%
  \institution{Vanderbilt University}
  \city{Nashville}
  \state{TN}
  \country{USA}}
\email{siyuan.ma@vumc.org}

\author{Jin Jin}
\authornote{Corresponding author.}
\affiliation{%
  \institution{University of Pennsylvania}
  \city{Philadelphia}
  \state{PA}
  \country{USA}}
\email{jin.jin@pennmedicine.upenn.edu}

\author{Lingyao Li}
\authornotemark[2]
\affiliation{%
  \institution{University of South Florida}
  \city{Tampa}
  \state{FL}
  \country{USA}}
\email{lingyaol@usf.edu}

\renewcommand{\shortauthors}{Duan and Wei, et al.}

\begin{abstract}
\textbf{Abtract:} Social media is a rich source of real-world data that captures valuable patient experience information for pharmacovigilance. However, mining data from unstructured and noisy social media content remains a challenging task. We present a systematic framework that leverages large language models (LLMs) to extract medication side effects from social media and organize them into a knowledge graph (KG). We apply this framework to semaglutide for weight loss using data from Reddit. Using the constructed knowledge graph, we perform comprehensive analyses to investigate reported side effects across different semaglutide brands over time. These findings are further validated through comparison with adverse events reported in the FAERS database, providing important patient-centered insights into semaglutide’s side effects that complement its safety profile and current knowledge base of semaglutide for both healthcare professionals and patients. Our work demonstrates the feasibility of using LLMs to transform social media data into structured KGs for pharmacovigilance.

\end{abstract}

\keywords{Pharmacovigilance, Large Language Models, Knowledge Graph, Social Media Mining, FAERS Database}

\maketitle

\section{Introduction}
When a medicine is approved, it typically undergoes continued monitoring for safety, including the detection of adverse drug reactions and long-term effects, a process known as pharmacovigilance \cite{lavertu2021new}. While randomized controlled trials (RCTs) are considered the gold standard for evaluating drug safety and efficacy, they have limitations due to strict inclusion and exclusion criteria, which could lead to the underrepresentation of certain patient populations commonly seen in clinical practice \cite{blonde2018interpretation}. In contrast, analyzing real-world data (RWD) from a broader and more diverse population over an extended period after a drug enters the market can help bridge the gap between trial results and patient outcomes in real-world settings to complement the evidence from RCTs \cite{lavertu2021new, blonde2018interpretation, wilson2024real, grimberg2021real, makady2017real}.

Traditional post-marketing surveillance relies on Phase IV trials and real-world evidence from electronic health records (EHRs), claims data, or the FDA Adverse Event Reporting System (FAERS) \cite{FDA2019RWE, FDA2020Postmarket}. In recent years, social media has gathered increasing attention as a potential source of RWD \cite{McDonald2019, Wessel2024, Sarker2015, Pappa2019}. Unlike the FAERS database, which releases adverse events (AEs) data quarterly \cite{FDA2023FAERS}, social media is characterized by real-time and dynamic user discussions \cite{li2022dynamic}, which could greatly facilitate the early detection of emerging safety signals. In addition, social media data is publicly accessible, which contrasts with EHRs that are restricted due to patient privacy and security concerns. Another important advantage of social media data is its patient-centered nature. Crowdsourcing this important source of personal experiences can bring valuable patient perspectives into drug post-marketing monitoring, which is essential to achieving better patient-centered care \cite{McDonald2019, Sarker2015, Pappa2019}.

However, social media data are inherently unstructured, scattered, and noisy, which poses challenges for information extraction and analysis. Previous studies have demonstrated the use of various tools---including natural language processing (NLP), thematic coding, and knowledge graphs (KGs)---to analyze user discussions on online platforms such as X, Reddit, and other forums \cite{Dirkson2023, Yu2022, Nikfarjam2015, Li2020}. The growing body of literature highlights the potential of these tools to identify off-label use and early signals of adverse effects of medications that are not readily captured in clinical settings \cite{Hua2022, Golder2021}. The mining of social media data can be further enhanced by emerging large language models (LLMs), which are well-suited for information extraction from large-scale, complex text content \cite{Xu2023LLM, xu2025patients}.

While studies have utilized social media data to analyze side effects and user experiences, few have focused on developing a comprehensive pipeline for constructing KGs from user-generated content to present structured information on real-world drug effects. Moreover, the feasibility of analyzing large-scale social media content and the potential of LLMs to enhance knowledge graph construction remain largely underexplored. We address these research gaps by developing a systematic pipeline for social media-based knowledge graph construction for drug side effects utilizing LLMs. A knowledge graph is composed of nodes and edges, with the basic unit of observation being a triple linking two objects through a relationship \cite{Hauben2024}. It offers the unique advantage of presenting complex drug-related information in a structured way. In particular, we investigate three research questions:

\begin{itemize}
    \item \textbf{RQ1}: How can social media data be transformed into a knowledge graph using LLMs to represent relationships between the medication and reported side effects?
    \item \textbf{RQ2}: What are the important messages on the major side effects identified through qualitative and quantitative analyses of the constructed knowledge graph?
    \item \textbf{RQ3}: How do the side effects identified from social media compare with the AEs reported in FAERS?
\end{itemize} 

We select semaglutide as the target medication to demonstrate the utility of our pipeline. Semaglutide, a glucagon-like peptide-1 (GLP-1) receptor agonist, was first approved by the FDA in December 2017 for the treatment of type 2 diabetes mellitus \cite{Dhillon2018}. In 2021, a new formulation received FDA approval for chronic obesity and overweight management, expanding its use to a broader population \cite{FDA2021Semaglutide}. According to the Centers for Disease Control and Prevention (CDC), approximately 40.3\% of U.S. adults were affected by obesity between August 2021 and August 2023 \cite{CDC2024Obesity}. Looking ahead, it is estimated that the total number of adults with overweight and obesity would reach 213 million by 2050 \cite{Ng2024}. This growing burdened population corresponds with the rising public interest in semaglutide, as reflected by the surge of online search queries and increasing off-label use in recent years \cite{Raubenheimer2024}. Investigating semaglutide through social media presents a timely opportunity for us to understand user perspectives and real-world side effects of medications beyond clinical trial settings.

Our contribution is twofold. First, we demonstrate how LLMs can be used to efficiently extract side effects information from unstructured social media data and construct an easy-to-use, publicly accessible knowledge base for downstream analysis and public references. Second, we present a scalable pipeline designed for LLM-assisted construction of KGs using social media data. We apply the pipeline to semaglutide, showcasing the value of crowdsourced social media data in complementing traditional pharmacovigilance and facilitating the generation of real-world evidence. Our case study highlights the applicability, validity, and efficiency of our pipeline for the general task of developing social media-based side effect KGs for various health conditions.

\section{Materials and Methods}

\begin{figure*}[htbp]
    \centering
    \includegraphics[width=1\textwidth]{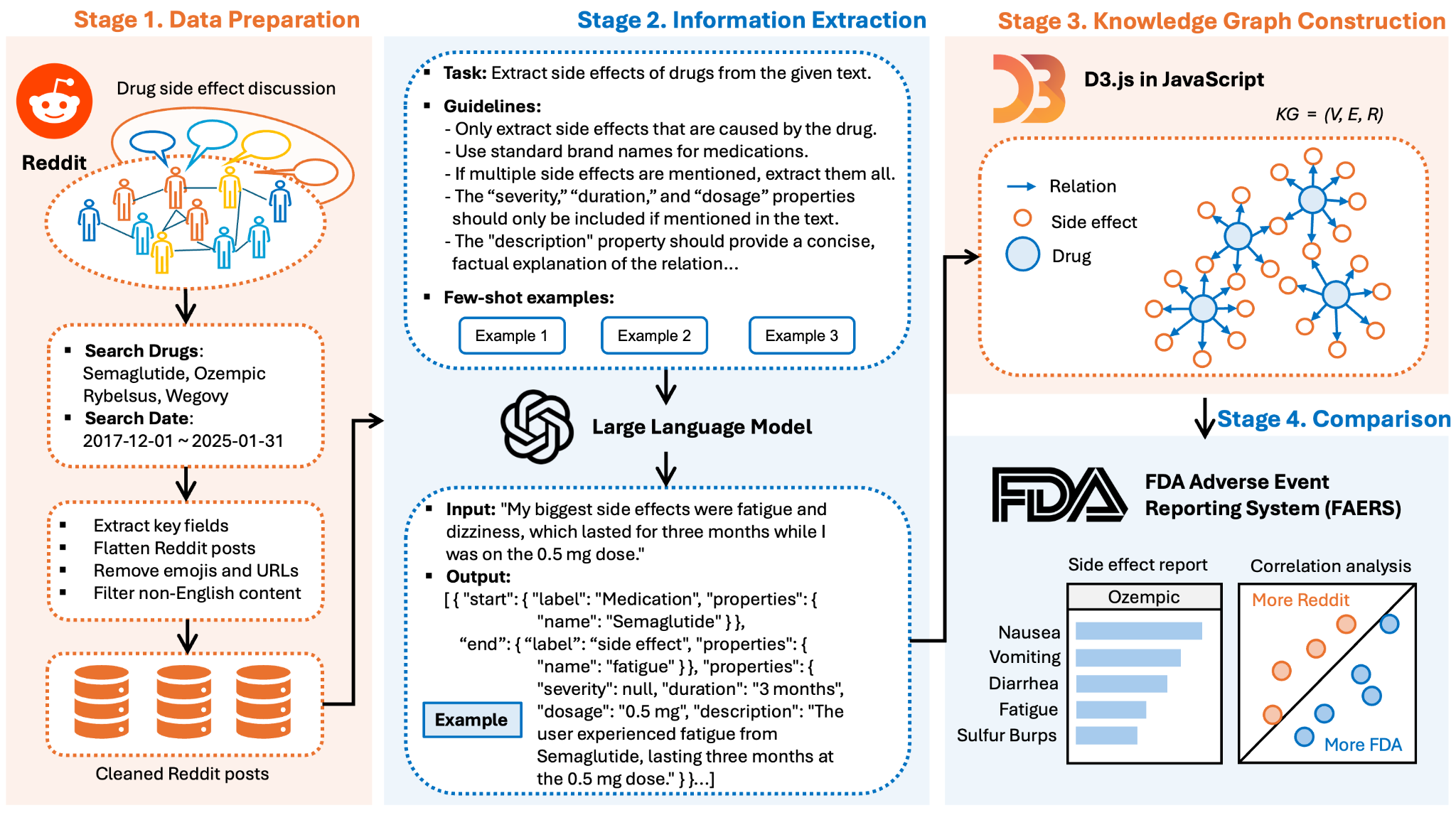} 
    \caption{The conceptualized framework of the pipeline applied to semaglutide.} 
    \label{fig1} 
\end{figure*}

The conceptual framework of the proposed pipeline is illustrated in Figure \ref{fig1} and consists of four stages. First, posts related to semaglutide and its brand names are retrieved from the Reddit platform, followed by data-cleaning steps such as the removal of emojis and non-English content. Second, a LLM, specifically GPT-4o-mini, is used to extract side effect information from the collected posts. Through prompt design and few-shot in-context learning, the model identifies side effects, their associated medications, and additional attributes such as severity, duration, and dosage when mentioned. Third, the extracted information is visualized using D3.js, where nodes represent drugs and side effects and edges denote the relationships between them. Finally, the Reddit-derived knowledge graph is compared with the FAERS database to highlight both overlaps and discrepancies, and a correlation analysis is performed to quantitatively assess the alignment between the two data sources.

\subsection{Data preparation}
We choose Reddit as our primary social media data source. Reddit is a popular platform where users can share content, which can be voted up or down by other users. Discussions are organized into user-created communities called subreddits, each focused on specific topics. A typical Reddit discussion consists of three parts: a post, which serves as a conversation starter; comments, where users engage with the post by leaving comments; and replies, where users can respond to comments, forming a hierarchical, tree-like structure. To collect relevant data, we conduct a search on February 1, 2025, for Reddit posts containing the keyword ``Semaglutide'' and its three brands, namely ``Ozempic,'' ``Rybelsus,'' and ``Wegovy.'' We used the Python-based library PRAW (Python Reddit API Wrapper) to download user-generated posts published between December 1, 2017, and January 31, 2025 \cite{Pickett2024, GoLogin2023Reddit}. The start date was chosen to align with Ozempic’s FDA approval on December 5, 2017 \cite{FDA2020Ozempic}.

We additionally collect data from FAERS, one of the largest pharmacovigilance databases in the world, which collects adverse event (AE) reports from pharmaceutical manufacturers, healthcare providers, patients, and other sources \cite{Yin2022, He2024}. FAERS uses preferred terms (PTs) from the Medical Dictionary for Regulatory Activities (MedDRA) to describe AEs \cite{Omar2021}. These PTs are organized into 27 system organ classes (SOCs), and each PT may be associated with multiple SOCs \cite{MedDRA2019}. We search the FAERS using four product names: ``Semaglutide'' (i.e., general semaglutide without brands specified), ``Ozempic,'' ``Rybelsus,'' and ``Wegovy,'' and separately download the data for each product. This includes all AEs associated with these products reported in FAERS up to the fourth quarter (Q4) of 2024.

After collection, the Reddit data undergoes several cleaning steps. First, we flatten the scraped data so that each observation in the transformed dataset represents either a post (conversation starter), a comment, or a reply. Each observation includes variables such as publication date, score (upvotes minus downvotes), ID, author, parent ID (linking it to the higher-level conversation), and text, among others. To further refine the dataset, we filter out observations with non-English text, bot-generated responses, or content consisting solely of a URL.

\subsection{Information extraction} 
In Stage 2 (Figure \ref{fig1}), we leverage LLM to perform information extraction, aiming to extract drug-relation-side effect triples related to semaglutide from the given text. We restrict the medication entity name to only ``Wegovy,'' ``Ozempic,'' and ``Rybelsus,'' as well as ``Semaglutide'' when no specific brand name is mentioned. For consistency, ``Semaglutide'' is later relabeled as ``Unspecified Brands'' in the subsequent analysis. Additionally, we instruct GPT to extract the ``severity,'' ``duration,'' and ``dosage'' of side effects if explicitly mentioned in the text and store them as properties of the relations. We also instruct GPT to provide a concise explanation of each relation within the properties. A prompt is a set of instructions provided to an LLM to influence its response generation and customize the output for our specific needs \cite{White2023, Gao2024}. We employ a few-shot prompting technique to enhance in-context learning (ICL) by including five examples in the prompt \cite{Gao2024}. Among the five examples, three are ``null'' examples, which demonstrate cases where no information should be extracted—either because the text does not align with the guidelines (e.g., discussing other drugs instead of semaglutide products) or because it lacks meaningful relations \cite{Wan2023}. For each observation, the extracted relations are saved into the variable ``Relations'' in the data. We select GPT-4o mini because it is a relatively lightweight model that offers a great balance of performance and cost efficiency \cite{OpenAI2024}. For the rest of the paper, we use ``GPT'' to refer to the GPT-4o mini model to maintain consistency and conciseness.

After extracting information, we remove observations with null values, as these posts, comments, or replies contain no extracted information. Next, we remove duplicates by keeping unique (ID, text) pairs, resulting in 4,242 rows. These rows contain 7,225 relations and 2,284 unique side-effect entities (end nodes). To improve the consistency of information extraction, we consolidate highly similar side effects into standardized terms. For example, ``Headaches'' and ``Headache'' are standardized as ``Headache,'' while variations like ``Liver Damage,'' ``Damaging Liver,'' and ``Damage to Liver'' are standardized as ``Liver Damage.'' This standardization is conducted in two steps. First, we use SentenceTransformer’s all-MiniLM-L6-v2 model, a pre-trained model designed for efficient semantic text similarity computation \cite{SBERT2025}, to generate embeddings for each unique side effect. We then construct a similarity matrix and filtered out entities with a similarity score below 0.9. This leaves us with 605 unique side effects requiring further grouping. Second, we leverage GPT to support clustering by iteratively grouping similar side effects. If GPT determines that a side effect matches an existing key, the side effect is added to the group of that key; otherwise, it becomes a new key. This two-step process results in a total of 96 side effects. Each group is manually reviewed and adjusted if necessary, with final names standardized using the PTs from MedDRA as the reference.

Next, we evaluate GPT’s performance in extracting side effects and their associated severity. We do not evaluate dosage or duration due to the high rate of missing data (only a few posts included that information). We randomly select 213 observations (5\% of the sample)  for evaluation. Each relation is independently evaluated by three authors who assign a score of 1 (correct) or 0 (incorrect) based on the criteria of faithfulness, completeness, and relevance \cite{Song2024}. The final annotation is determined through a majority-voting strategy \cite{Wang2024}. The final accuracy of the extracted side effects and their associated severity, as determined through manual annotation, is 79.34\% and 86.85\%, respectively.

\begin{table*}[htbp]

\caption{Representative examples of side effect extraction using GPT-4o mini.}
\label{tab:side_effects}
\centering
\begin{tabularx}{\textwidth}{p{5cm} p{3.5cm} p{1.8cm} p{1cm} p{1cm}
}
\toprule
\textbf{Original Post} & \textbf{Extracted Relations} & \textbf{Side Effects} & \textbf{Severity} & \textbf{Dosage} \\
\midrule
11 weeks in on my Ozempic. Have had 2 doses at 1mg injected into thigh which is almost painless. I alternate thighs each week. At .25 I would have severe cramping in stomach the day after injection. Eating carbs made it worse.&
\scriptsize\ttfamily 
[{"start": {"label": "Medication", "properties": {"name": "Ozempic"}}, "end": {"label": "SideEffect", "properties": {"name": "Stomach Cramps"}}, "properties": {"severity": "severe", "duration": "1 day", "dosage": "0.25 mg", "description": "The user experienced severe cramping in the stomach the day after the 0.25 mg injection."}}]
 &
Stomach Cramps &
Severe & 0.25 mg \\

I've been fine. I have been on a low dose though as that's been enough. I had very mild nausea on 1.2mg, I think because I didn't actually eat enough. It worked too well. My appetite is reduced enough on 0.6mg so I've gone back down to that.&
\scriptsize\ttfamily 
[{"start": {"label": "Medication", "properties": {"name": "Semaglutide"}}, "end": {"label": "SideEffect", "properties": {"name": "Nausea"}}, "properties": {"severity": "mild", "duration": null, "dosage": "1.2 mg", "description": "The user experienced mild nausea from semaglutide at the 1.2 mg dose."}}]&
Nausea &
Mild & 1.2 mg \\

I was on saxenda 3.0 for a year, then went to .25 on ozempic, week two went to .5, regretted it due to depressed mood, did three more weeks on .25, three weeks on .5, and am on week 3 of 1.0. I didn't really lose anything until 1.0, but I wasn't losing on saxenda for the last several months so it was no skin off my back to titrate up.&
\scriptsize\ttfamily 
[{"start": {"label": "Medication", "properties": {"name": "Ozempic"}}, "end": {"label": "SideEffect", "properties": {"name": "Depression"}}, "properties": {"severity": "severe", "duration": null, "dosage": "0.5 mg", "description": "The user experienced severe side effects after increasing the dosage of Ozempic to 0.5 mg."}}]&
Depression &
Severe & 0.5 mg \\
\bottomrule
\end{tabularx}
\end{table*}

\subsection{Knowledge graph construction}
Using the cleaned data, we construct a systematic KG structure to represent the relationships between semaglutide products and their reported side effects, as presented in Figure \ref{fig1} (Stage 3). The KG can be formally defined as:

\[
\text{KG} = (V, E) \tag{1}
\]

where \( V \), \( E \) denote the set of vertices (nodes) and edges (relationships), respectively. In our semaglutide side effect KG, each node belongs to one of two categories:

\[
V = M \cup S \tag{2}
\]

where \( M = \{m_1, m_2, \ldots, m_n\} \) represents medication entities (Wegovy, Ozempic, Rybelsus, and the generic term "Unspecified Brands"), and \( S = \{s_1, s_2, \ldots, s_n\} \) represents the side effects entities. The relationships between medications and side effects are represented as directed edges, where:

\[
e = (m, s, P) \tag{3}
\]

Here, \( m \in M \) is a medication entity, \( s \in S \) is a side effect entity, and \( P = \{p_1, p_2, \ldots, p_n\} \) is a set of properties that capture additional metadata, including:

\[
P = \{\text{severity}, \text{duration}, \text{dosage}, \text{description}\} \tag{4}
\]

These properties are populated when available in the source content, allowing for rich contextual information beyond simple links.

To deploy the constructed KG, we utilize D3.js, a JavaScript visualization library that enables interactive data representation. We employ a force-directed graph layout algorithm that positions nodes optimally by simulating physical forces, where nodes repel each other while connected nodes are pulled together, resulting in a visually intuitive arrangement that minimizes edge crossings. Our visualization assigns distinct visual encodings to medication and side effect nodes. The radius of each side effect node is proportional to its frequency of mention in the dataset, with node sizes calculated as:

\[
\text{radius}(s) = \text{base\_size} + \log(\text{frequency}(s)) \tag{5}
\]

Similarly, edge thickness is scaled according to the weight of the relationship, reflecting how frequently a particular medication-side effect pair co-occurred in the Reddit corpus:

\[
\text{thickness}(e) = \text{base\_thickness} \cdot \log(\text{weight}(e)) \tag{6}
\]

\subsection{Comparison between FDA adverse events and Reddit-extracted side effects}
In Stage 4, we compare Reddit-derived side effects with FDA-reported AEs. By manually matching the top 20 side effects extracted from Reddit with the top 20 adverse events reported in the FDA FAERS database (Figure \ref{fig4}A), we calculate the frequency of each side effect/AE pair in either FDA reports or Reddit posts. For FDA-reported AEs, the frequency for an AE \( j \) is defined as:

\[
\text{Freq}_j^{\text{FDA}} = \frac{\# \text{Adverse Event } j}{\# \text{Total FAERS Reports}} \tag{7}
\]

where the number of total reports accounts for all patient reports on AEs related to Semaglutide. Similarly, for the same side effect \( j \) as reported by Reddit, its frequency is calculated as:

\[
\text{Freq}_j^{\text{Reddit}} = \frac{\# \text{Side Effect } j}{\# \text{Total Reddit Posts}} \tag{8}
\]

where the number of total Reddit posts, by design of our study, represents the number of posts reporting any side effects of Semaglutide. Thus, while neither metric should be interpreted as the incidence rate of AEs/side effects in the general population, they do represent similar statistics, namely, the propensity for a particular AE or side effect for Semaglutide, among all such reports as collected by the FDA or via Reddit crowdsourcing (Figure \ref{fig4}B).

We conduct a formal binomial regression analysis to test whether the difference between Reddit-based and FDA-based frequencies is statistically significant. Specifically, for a specific side effect/AE pair \( j \), we ran the following regression:

\[
\# \text{Side Effect}_j, \# \text{Total Posts}_j \sim \beta_0 + \beta_1 X_j \tag{9}
\]

where \( X_j \) is the binary covariate indicating either FDA-reported adverse events (\( X_j = 0 \)) or Reddit side effects (\( X_j = 1 \)). The above formulation thus exactly represents the frequency metrics as defined above but allows for formal statistical tests for whether the difference between Reddit versus FDA (\( \beta_1 \); log odds ratio for side effect propensity in Reddit compared to FDA) is statistically significant, as reported in Figure \ref{fig4}C. P-values across comparisons for each adverse event are corrected with the Bonferroni procedure.

Lastly, as the FAERS database reports adverse events as associated with specific Semaglutide brands (Ozempic/Wegovy/Rybelsus), we further conduct the above analyses but are restricted to each individual brand (Figure \ref{fig4}D).

\section{Results}
\subsection{RQ1. Knowledge graph for semaglutide side effects}

\begin{figure*}[htbp]
    \centering
    \includegraphics[width=0.9\textwidth]{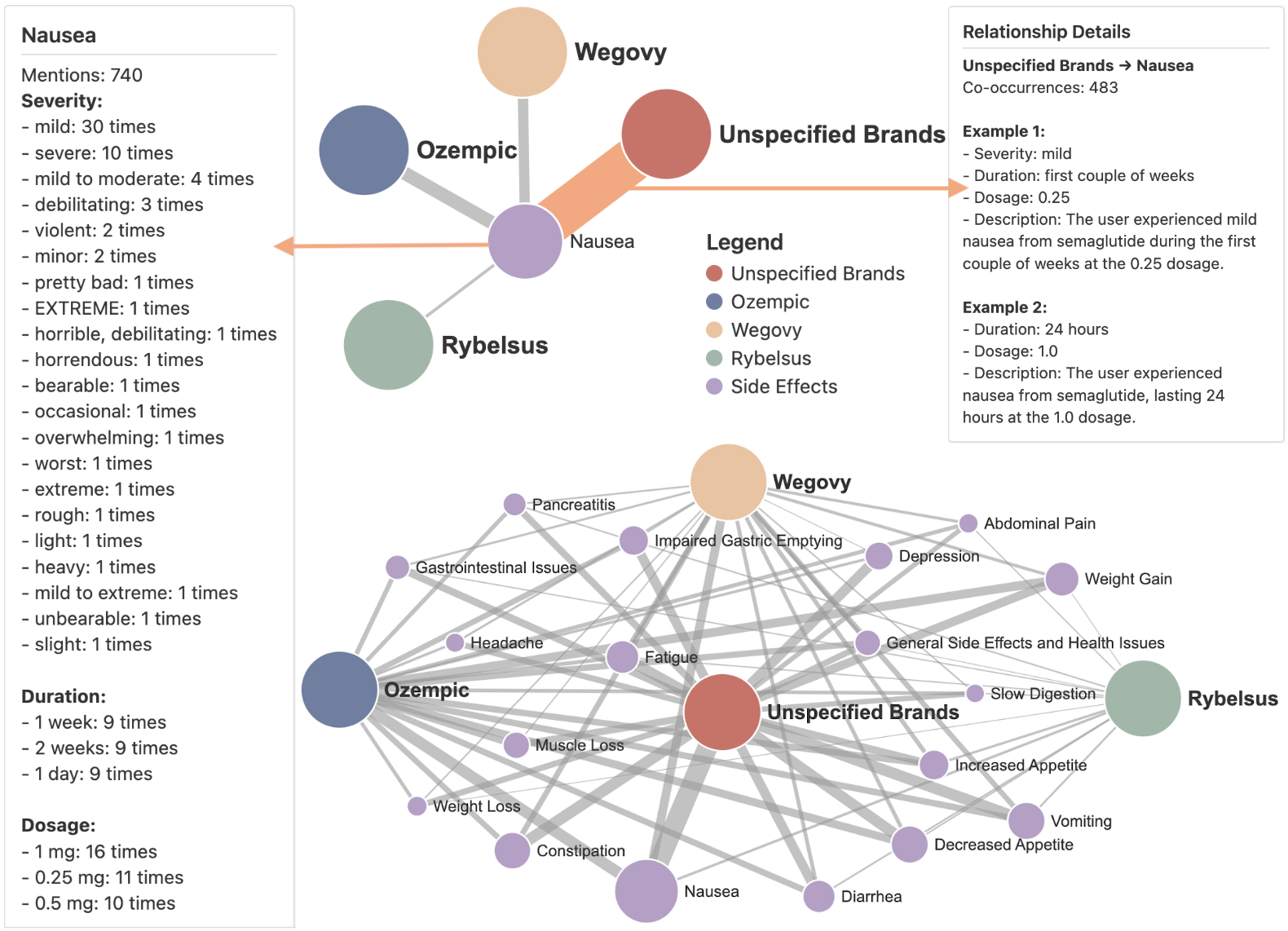} 
    \caption{Interactive knowledge graph visualization of semaglutide side effect information extracted from Reddit data. \textbf{A)} The knowledge graph displaying relationships between four medication entities and their associated 
side effects, with node sizes proportional to the frequency of Reddit mentions. \textbf{B)} Focused view of the knowledge 
graph when selecting nausea as a single side effect, showing detailed statistics on severity, duration, and dosage, along 
with representative user experiences across different medication brands.} 
    \label{fig2} 
\end{figure*}

The knowledge graph constructed from GPT-extracted relations visualizes the complex interaction network between semaglutide products and their associated side effects reported on Reddit (Figure \ref{fig2}). The graph represents 7,225 relations connecting four medication brand entities (Ozempic, Wegovy, Rybelsus, and unspecified brands) with 1,775 unique side effect entities. To distinguish entity types, we assign different colors to medication nodes and side effect nodes, with side effect nodes sized proportionally to their mention frequency. Analysis of the graph reveals that nausea is the most frequently reported side effect, which is visually represented by its significantly larger node size than other side effects. The connection between unspecified brands and nausea appears substantially thicker than connections to less frequently reported side effects, reflecting the weight of this relationship in user discussions. 

The interactive capabilities of our implementation enable a closer look and more detailed examination of specific elements on the graph. Figure \ref{fig2}B demonstrates this functionality, where selecting ``Nausea'' as a single side effect automatically generates a focused knowledge graph displaying the connections between all four medication brand entities and the nausea node. When hovering over a side effect node such as ``Nausea'' (Figure \ref{fig2}B), a tooltip displays comprehensive statistics, including 740 total mentions with severity distributions ranging from ``mild'' (30 mentions) to ``severe'' (10 mentions) and various other descriptors. Duration data reveals equal distribution among common reporting periods (e.g., ``1 week,'' ``2 weeks,'' and ``1 day,'' each with 9 mentions), while dosage information indicates variation across medication levels with ``1 mg'' being the most frequently mentioned (16 instances). Edge tooltips provide detailed insights into specific medication-side effect relationships (Figure \ref{fig2}B). The Ozempic-nausea relationship displays 483 co-occurrences, along with representative examples that contextualize user experiences. For instance, the tooltip shows two patient examples with varying experiences: one user reported nausea for the first couple of weeks at a 0.5 unit dosage, while another experienced nausea for 24 hours at a higher 1.0 unit dose. Our interactive KG is publicly accessible at 
\url{https://zx136.georgetown.domains/Semaglutide/knowledge_graph.html}.

\subsection{RQ2. Qualitative and quantitative analysis of semaglutide side effects}

\begin{figure*}[htbp]
    \centering
    \includegraphics[width=1\textwidth]{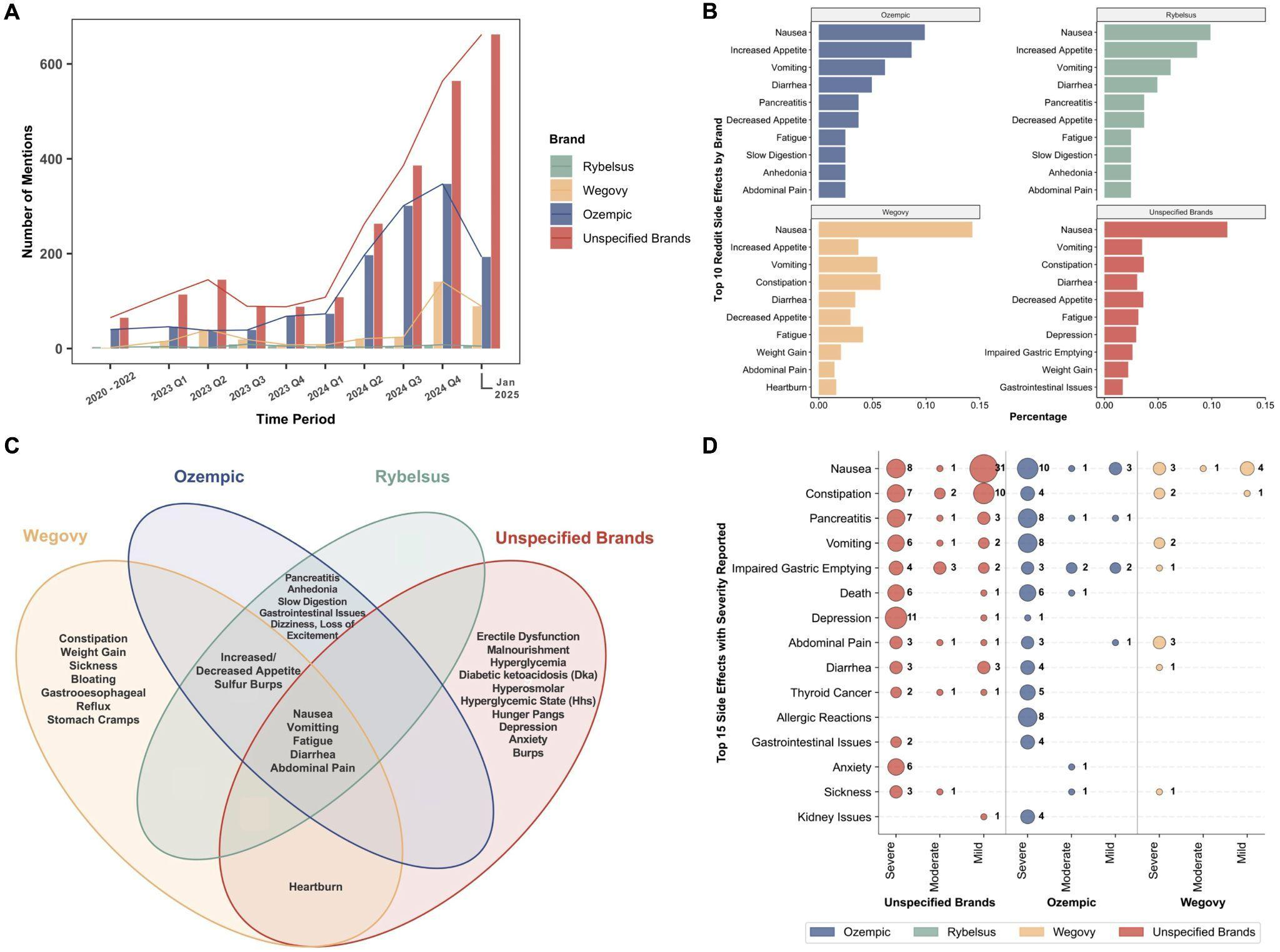} 
    \caption{An overview of GPT-extracted Reddit post information on semaglutide. \textbf{A)} Number of Reddit mentions of semaglutide side effects by brand between January 1st, 2020, and January 31st, 2025. \textbf{B)} Top 10 GPT-extracted side effects mentioned for each brand of semaglutide. \textbf{C)} Overlapping top side effects extracted across brands. \textbf{D)} Top 15 GPT-extracted side effects weighted by GPT-extracted severity across brands. The point size indicates the number of mentions. Rybelsus is not displayed in the figure because none of the top 15 side effects with comments on severity were mentioned for Rybelsus.} 
    \label{fig3} 
\end{figure*}

We first examine the GPT-extracted Reddit post information on semaglutide, including side effects and their severity/dosage information (if available), across different brands and over time (Figure \ref{fig3}). We observe that very few semaglutide-related side effects are discussed on Reddit before 2022, with only 14 mentions in 2020 and 65 in 2021. However, discussions have increased sharply since 2023, with 1,215 mentions in 2023 and 4,195 mentions in 2024, and monthly mentions peaking in the most recent month (1,614 relations in Jan. 2025, Figure \ref{fig3}A). This reflects the growing popularity of all brands of Semaglutide, including Ozempic, Wegovy, Rybelsus, and emerging new brands, for weight loss, as public interest and media coverage surged in recent years \cite{Watanabe2024}. The decrease in 2025 in Figure \ref{fig3}A is due to our data collection ending on January 31, 2025, resulting in only one month of data in the first quarter of 2025. The overall upward trends for Ozempic, Wegovy, and unspecified brands suggest the increasing popularity of these brands in the following years. On the other hand, the number of mentions per quarter for Rybelsus never exceeds 9 and does not show a clear increasing or decreasing pattern over the years, suggesting that Rybelsus may not be as popular and accepted as the other brands over the past five years.

The top 10 side effects extracted from Reddit across all brands are nausea (14.3\%), decreased appetite (3.8\%), weight gain (3.2\%), impaired gastric emptying (3.2\%), constipation (2.7\%), depression (2.5\%), increased appetite (2.4\%), fatigue (2.4\%), muscle loss (2.2\%), and vomiting (2.2\%). The top side effects overlap considerably but also vary across different brands of Semaglutide (Figure \ref{fig3}B). Among the top 20 side effects of each brand, the five top side effects shared by all brands are mostly gastrointestinal symptoms (with one exception, ``fatigue''), while other unique and shared side effects are also gastrointestinal issues mainly, with other symptoms like negative emotion and depressive symptoms (all but Wegovy) and heartburn (Wegovy and unspecified brands). Finally, we examine the extracted severity information associated with the top 15 side effects across all brands and observe that most of the comments on severity levels focus on ``severe'' followed by ``mild'' (Figure \ref{fig3}D). Among the three brands, Ozempic shows the highest frequency of ``severe'' side effects, particularly nausea, vomiting, pancreatitis, and allergic reactions. Wegovy’s profile shows intermediate reporting rates, with nausea and abdominal pain being the most frequent severe side effects. Unspecified brands demonstrate notably high reports of nausea, constipation, and pancreatitis, but a large proportion of them only show mild side effects.

\subsection{RQ3. Validation via comparison with adverse events reported in FAERS}

\begin{figure*}[htbp]
    \centering
    \includegraphics[width=1\textwidth]{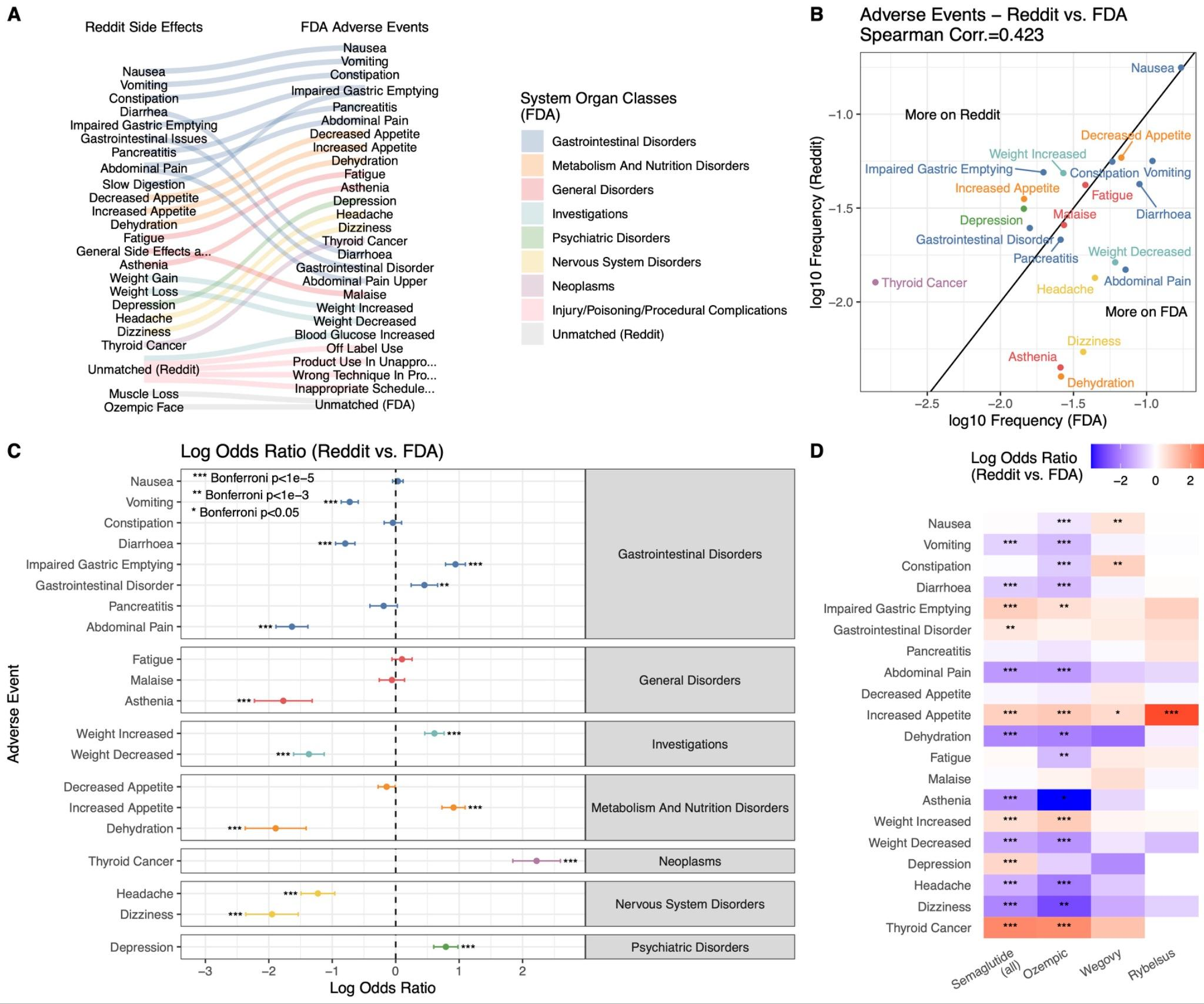} 
    \caption{Crowd-sourced side effect surveillance based on Reddit posts for semaglutide demonstrates mostly consistent findings with FDA-registered AEs, with additional unique discoveries. \textbf{A)} Correspondence 
between the top 20 side effects identified by our Reddit-based knowledge graph and top AEs registered by the FDA. 
\textbf{B)} Frequency of top events as measured by the FDA compared to Reddit posts largely also agree with each other 
(Spearman correlation of per-adverse-event frequencies=0.423). \textbf{C)} Binomial logistic regression formally established 
events that Reddit crowdsourcing differentially detected compared to FDA (log odds ratio; statistical significance 
determined with p<0.05 post-Bonferroni correction across AEs). \textbf{D)} Brand-specific analysis of the differential pattern 
of Reddit crowdsourcing versus FDA surveillance.} 
    \label{fig4} 
\end{figure*}

Top side effects as identified by our Reddit-based knowledge graph demonstrate agreement with  FDA-registered top adverse events (AEs) (Figure \ref{fig4}). Most of the top twenty Reddit side effects are successfully matched against top AEs recorded at the FDA FAERS (Methods, Figure \ref{fig4}A) The only notable exceptions are (1) Reddit posts fail to cover adversarial procedural complications noted by the FDA (e.g., off-label use, product use in unapproved indication) and (2) Reddit posts highlight ``minor'' side effects—such as muscle loss and the so-called ``Ozempic Face''—that the FDA does not officially classify as adverse events. In addition to nominal matches, the frequencies of these top events are also positively correlated between the two platforms (Spearman correlation=0.423; Figure \ref{fig4}B)). Frequency of FDA AEs or Reddit side effects is calculated as the prevalence of an event among all event reports or Reddit posts and represents the propensity for that event among all cases under either platform (Methods). Our findings thus support concordance in Semaglutide usage’s most common side effects, thus validating Reddit crowdsourcing-based results against established FDA records (RQ3). 

Beyond this overall agreement, Reddit crowdsourcing also identifies nuanced differences against FDA reports (Figure \ref{fig4}B-D). Among AEs that are differentially detected between the two approaches, crowd-sourcing demonstrates a higher tendency to report rarer AEs under FDA surveillance (impaired gastric emptying, general gastrointestinal disorders, increased appetite weight increase, thyroid cancer, and depression; Figure \ref{fig4}B, C). Statistical significance is established with formal binomial regression and corrected for multiple testing (Methods). These findings, particularly those related to the treatment’s less expected outcomes (e.g., increased appetite and weight gain), suggest that Reddit users tend to discuss side effects that are less commonly recorded by the FDA. Lastly, in medication brand-specific analysis (Methods), the differential pattern of Reddit crowdsourcing versus FDA surveillance largely remains consistent across medication brands (Figure \ref{fig4}D). Most prominently, increased appetite is significantly more frequent in Reddit posts compared to FDA across all brands. Qualitative differences were only observed for two gastrointestinal conditions, nausea and constipation, which were detected more frequently under Wegovy usage among Reddit posts than FDA but less frequently under Ozempic usage. Such secondary differences between our crowdsourcing-identified side effects in comparison to FDA reports highlight the complementary capacity of crowdsourcing relative to official registrant data, in particular for events less commonly registered in the latter (RQ3).

\section{Discussion and Conclusions}
We use semaglutide as a case study to introduce a structured pipeline that utilizes LLMs to extract and process information on drug side effects from large social media platforms and formulates it into a comprehensive KG. This framework supports future pharmacovigilance by integrating social media as a supplementary source of RWD.

To address RQ1 (see Introduction), we develop a multi-stage pipeline that extracts side effect relations from Reddit posts using LLM and then organizes them into a knowledge graph linking drugs, side effects, and their associated properties. Compared to earlier NLP methods based on keyword matching \cite{sellami2022keyword} or co-occurrence frequencies \cite{onan2016ensemble}, our approach captures more nuanced language and semantic relationships, significantly enhancing recall and contextual relevance. In addition, this pipeline can process crowdsourcing data from social media, which captures real-world drug use and user-centered experience that often remain underrepresented in conventional clinical datasets, such as depression caused by semaglutide in our results. This insight can reveal latent patient concerns and subtle drug interactions that could be valuable for promoting pharmacovigilance strategies. 

Regarding RQ2 and RQ3 (see Introduction), analysis of the knowledge graph shows that the most common side effects are largely similar across different brands of Semaglutide, but there are also some differences. All three brands share the same top seven side effects, most of which are gastrointestinal issues (e.g., nausea and vomiting). This is consistent with reports from the FAERS database. Among the three brands, Ozempic has the highest number of reports of severe side effects. We also find that mental health-related symptoms—such as depression, anxiety, and mood swings—appear more often than FAERS. These symptoms are usually described in user posts that mention emotional struggles or a lack of motivation, which are often missed or underreported in traditional adverse event reporting systems. These findings also suggest that psychological symptoms, though frequently overlooked or underreported in traditional AEs reporting systems, may play a significant role in shaping patient experiences for weight loss drug use. This further underscores the value of patient-shared experiences in capturing a broader and more nuanced spectrum of adverse effects. In addition, Reddit posts often appear to surface less common or emerging concerns, such as thyroid cancer, showing the platform's potential for identifying side effects that may be underreported in existing databases.

Our pipeline also has room for growth. First, we make a trade-off in this study by selecting GPT-4o mini as the backbone model for information extraction to ensure cost-effectiveness. While the model demonstrates strong performance, it occasionally misinterprets prompts—for instance, including side effects related to other drugs like Zepbound. Future work could explore alternative models and refined prompt engineering to improve accuracy.

Second, there is a conceptual challenge in distinguishing AEs from side effects. While AEs encompass all undesirable experiences during drug use, side effects imply a confirmed causal link. However, this nuance is often lost in public discourse; users on social media commonly use ``side effect''  as a catch-all. To align with user language, we use ``side effect'' during extraction but compared results with FDA-reported AEs for rigor. Future research can focus on better integrating official and user-generated data sources.

Despite these limitations, this work contributes to advancing patient-centered pharmacovigilance by demonstrating the potential of combining user-generated online content and rapidly evolving AI tools to capture patient voices.


\bibliographystyle{IEEEtran}
\bibliography{references.bib}

\appendix

\section{Prompt design}
\label{sec:prompt}
\begin{lstlisting}[language=Python]
PROMPT = {
    "Task": (
      "Extract relations about the side effects of semaglutide, including Wegovy, Ozempic, and Rybelsus, or semaglutide in general, from the given text."
      "Here is the given text: {text}"
    ),

    "Instruction": """ 
    Return the extracted relations in the exact format below:

    [
      {
        "start": {
            "label": "Medication",
            "properties": {"name": ""}},
        "end": {
            "label": "SideEffect",
            "properties": {"name": ""}},
        "properties": {
            "severity": null,
            "duration": null,
            "dosage": null,
            "description": ""}
      },
      ...
    ]

    ### Guidelines:
    - **Only extract side effects that are caused by semaglutide.** 
      - **Do NOT extract side effects caused by other drugs (e.g. Mounjaro(tirzepatide)).**
    - **Use standard brand name** for medications:  
      - "Wegovy", "Ozempic", "Rybelsus"  
    - **If no specific brand name is mentioned**, use **"Semaglutide"** as the medication name.  
    - **If multiple side effects are mentioned**, extract them all in **separate relation entries**.  
    - **The "severity", "duration", and "dosage" properties should only be included if explicitly mentioned in the text.**  
      - If severity, duration, or dosage is not stated, set them to `null`.
    - **The "description" property** should provide a **concise, factual explanation** of the relation.  
    - **If no relevant relations can be extracted, return `null`.** 
    - **Do NOT include any explanations or additional information beyond the specified format.**  
    - **Return ONLY valid JSON.** Do NOT include `"json"` or any other prefix in your response.


    ### Example 1:
    **Input Text:**  
    "My biggest side effects were fatigue and dizziness, which lasted for three months while I was on the 0.5 mg dose."
    **Expected Output:**  
    [
      {
        "start": {
            "label": "Medication",
            "properties": {"name": "Semaglutide"}},
        "end": {
            "label": "SideEffect",
            "properties": {"name": "fatigue"}},
        "properties": {
            "severity": null,
            "duration": "3 months",
            "dosage": "0.5 mg",
            "description": "The user experienced fatigue from semaglutide, lasting three months at the 0.5 mg dose."}},
      {
        "start": {
            "label": "Medication",
            "properties": {"name": "Semaglutide"}},
        "end": {
            "label": "SideEffect",
            "properties": {"name": "dizziness"}},
        "properties": {
            "severity": null,
            "duration": "3 months",
            "dosage": "0.5 mg",
            "description": "The user experienced dizziness from semaglutide, lasting three months at the 0.5 mg dose."}}
    ]
    
    ### Example 2:
    **Input Text:**  
    "It also makes you have some severe stomach pain if you overeat anyway."
    **Expected Output:**
    [
      {
        "start": {
            "label": "Medication",
            "properties": {"name": "Semaglutide"}},
        "end": {
            "label": "SideEffect",
            "properties": {"name": "stomach pain"}},
        "properties": {
            "severity": "severe",
            "duration": null,
            "dosage": null,
            "description": "Severe stomach pain occurs if you overeat."}
      }
    ]        
    
    ### Example 3:
    **Input Text:**  
    "I've been using sema for months and had no side effects."
    **Expected Output:**  
    null
    **Return null if the text states that no side effects were experienced**  
    
    ### Example 4:
    **Input Text:**  
    "Ozempic does not cause headaches."
    **Expected Output:**  
    null
    **Return null if the text states that semaglutide does NOT cause a side effect**
    
    ### Example 5:
    **Input Text:**  
    "I experienced headaches after taking Tirzepatide."
    **Expected Output:**  
    null
    **Return null if the text is talking about another drug other than semaglutide**
    """
    }
\end{lstlisting}
\end{document}